\documentclass[10pt,twocolumn,letterpaper]{article}

\usepackage[pagenumbers]{cvpr} 

\usepackage[accsupp]{axessibility}  
\usepackage{graphicx}
\usepackage{amsmath}
\usepackage{amssymb}
\usepackage{booktabs}

\usepackage{xcolor}
\usepackage{bbm}

\usepackage[skip=0pt]{caption}
\usepackage[pagebackref,breaklinks,colorlinks]{hyperref}

\usepackage[capitalize]{cleveref}
\crefname{section}{Sec.}{Secs.}
\Crefname{section}{Section}{Sections}
\Crefname{table}{Table}{Tables}
\crefname{table}{Tab.}{Tabs.}

\def\cvprPaperID{1869} 
\def\confName{CVPR}
\def\confYear{2022}

\begin{document}

\title{AUV-Net: Learning Aligned UV Maps for Texture Transfer and Synthesis}
\author{ Zhiqin Chen$^{1,2}$  \quad Kangxue Yin$^{1}$ \quad Sanja Fidler$^{1,3,4}$ \\
	\quad \small{NVIDIA\textsuperscript{1}  \quad Simon Fraser University\textsuperscript{2} \quad University of Toronto\textsuperscript{3} \quad Vector Institute\textsuperscript{4}} 
}

\maketitle

\begin{abstract}

In this paper, we address the problem of texture representation for 3D shapes for the challenging and underexplored
tasks of  texture transfer and synthesis.
Previous works either apply spherical texture maps 
which may lead to large distortions, or use continuous texture fields that yield smooth outputs lacking details. 
We argue that the traditional way of representing textures with images and linking them to a 3D mesh via UV mapping is more desirable, since synthesizing 2D images is a well-studied problem.
We propose \emph{AUV-Net} which learns to embed 3D surfaces into a 2D aligned UV space, by mapping the corresponding semantic parts of different 3D shapes to the same location in the UV space. 
As a result, textures are aligned across objects, and can thus be easily synthesized by generative models of images.
Texture alignment is learned in an unsupervised manner by a simple yet effective texture alignment module, taking inspiration from traditional works on linear subspace learning.
The learned UV mapping and aligned texture representations enable a variety of applications including texture transfer, texture synthesis, and textured single view 3D reconstruction. We conduct experiments on multiple datasets to demonstrate the effectiveness of our method.
Project page: \href{https://nv-tlabs.github.io/AUV-NET}{https://nv-tlabs.github.io/AUV-NET}.

\end{abstract}

\vspace{-3mm}
\section{Introduction}
\label{sec:intro}
\vspace{-1mm}

\begin{figure}[t!]
\vspace{-2.5mm}
\begin{center}
\includegraphics[width=1.0\linewidth]{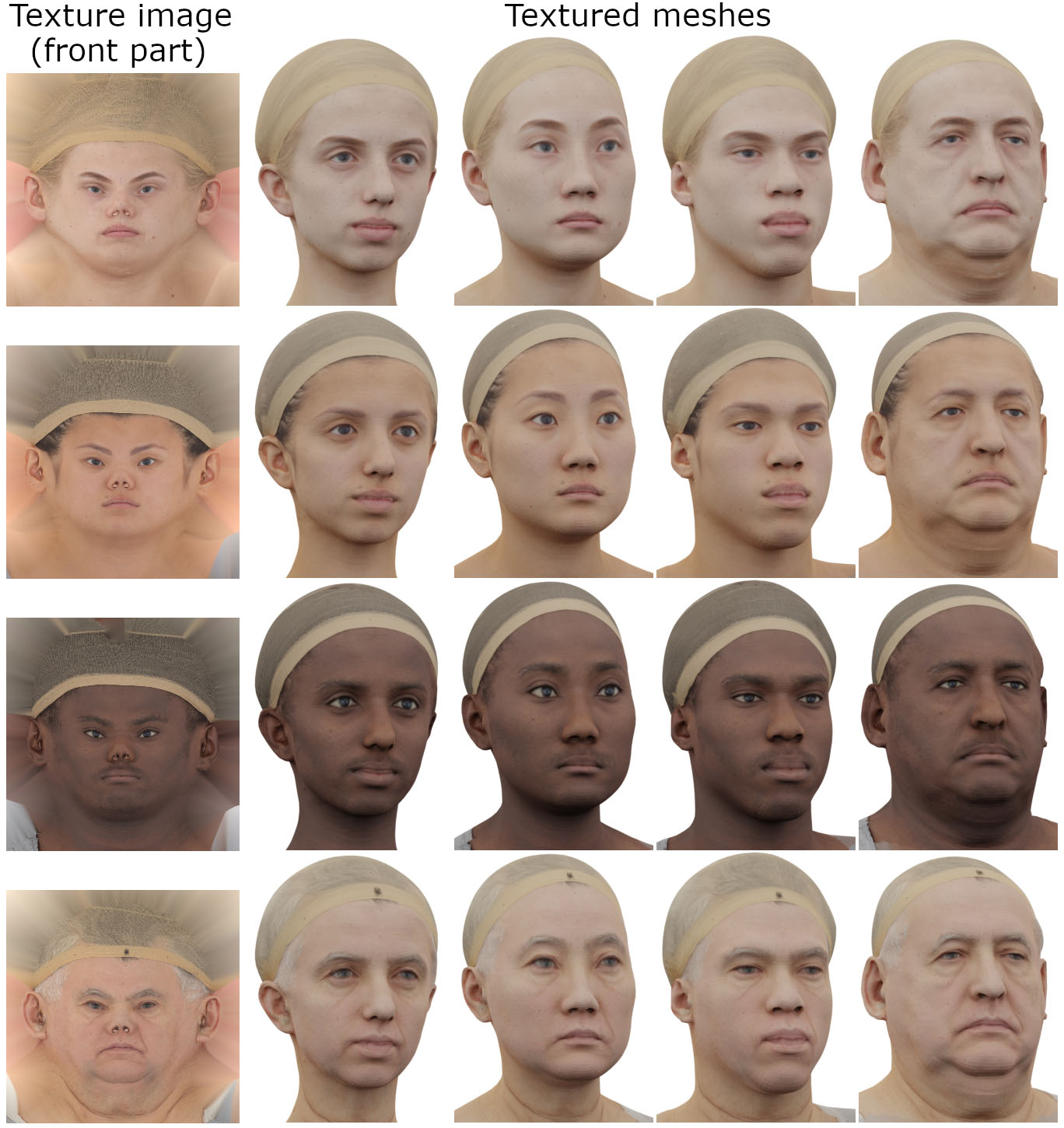}
\end{center}
\caption{
AUV-Net learns aligned UV maps for a set of 3D shapes, enabling us to easily transfer textures between shapes.
}
\label{fig:teaser}
\end{figure}

The field of 3D shape reconstruction and synthesis has witnessed significant advancements in the past few years. 
By utilizing the power of deep learning, several works reconstruct 3D shapes from voxels, point clouds, single and multi-view images, with a variety of output shape representations~\cite{choy20163d,AtlasNet,imnet,mescheder2019occupancy,style3d,kaolin,chen2020bspnet}. 3D generative models have also been proposed to synthesize new shapes ~\cite{PCGAN,DeepSDF,gao2019sdm,decorgan,dmtet21}, with the aim of democratizing 3D content creation. However, despite the importance of textures in bringing 3D shapes to life, very few methods have tackled semantic-aware texture transfer or synthesis for 3D shapes
~\cite{TMNET,pavllo2020convolutional,bhattad2021view,yin2021_3DStyleNet,henderson20cvpr,dibr,Raj_2019_CVPR_Workshops}.

Previous work on texture generation mostly relies on warping a spherical mesh template to the target shape ~\cite{pavllo2020convolutional,bhattad2021view,henderson20cvpr,dibr}, therefore obtaining a texture map defined on the sphere's surface, which can be re-projected into a square image for the goal of texture synthesis. 
NeuTex~\cite{xiang2021neutex} generates 3D shapes with a neural implicit representation for arbitrary surface topology, yet embeds the surface of the shape onto a sphere, which also results in a spherical texture map. Spherical texture maps can only support limited topology, and may introduce severe distortions for thin parts such as animal limbs ~\cite{tulsiani2020implicit,vmr2020}.
Another line of work uses implicit texture fields for texture synthesis~\cite{texture_fields}, without relying on explicit texture mapping. Although texture fields were successfully applied to multi-view image reconstruction~\cite{mildenhall2020nerf}, they have primarily been used for 
fitting a single object or scene. Generative models usually suffer from overly smoothed synthesized textures~\cite{schwarz2020graf,yenamandra2021i3dmm}.

In contrast, the traditional UV mapping in computer graphics handles arbitrary shape topology and avoids heavy distortions by cutting the surface into pieces and mapping different pieces to different regions on the 2D UV plane.
It further preserves texture details by storing the texture in a high-resolution texture image.
However, the UV mappings are usually created by 3D artists, and thus are inconsistent across different shapes. Therefore, using such representation for texture synthesis and transfer would require dense shape correspondences.

In this paper, we propose to train a neural network to predict the UV mapping and the texture image jointly, aiming at high-quality texture transfer and synthesis without needing to conform to a pre-defined shape topology. Specifically, our network learns to embed 3D coordinates on mesh surfaces into a 2D aligned UV space, where corresponding parts of different 3D shapes are mapped to the same locations in the texture image, as shown in Figure~\ref{fig:teaser}.
Such alignment is enabled by a simple yet effective texture alignment module inspired by traditional linear subspace learning methods such as Principal Component Analysis (PCA), as shown in Figure~\ref{fig:toy_net}.
The network generates a basis shared by all shape textures, and predicts input-specific coefficients to construct the texture image for each shape as a linear combination of the basis images.
This forces the texture images to be aligned so that they can be effectively decomposed into combinations of basis images, as visualized in Figure~\ref{fig:toy_result}.
Afterwards, the network reconstructs the colors of the input shape by learning a UV mapping to index the aligned texture image.
To unwrap 3D shapes of complex structure or topology, we further introduce a masking network that cuts the shape into multiple pieces to reduce the distortion in the UV mapping.

Our method effectively aligns textures across all shapes, allowing us to swap textures between different objects, by simply replacing the texture image from one object with another.
The aligned high-quality texture images produced by our method make it significantly easier to train generative models of textures, since they are aligned and disentangled from geometry. They also enable textured 3D shape reconstruction from single images.
We perform extensive experiments on multiple categories including human heads, human bodies, mammals, cars, and chairs, to demonstrate the efficacy of our approach.

\vspace{-2mm}
\section{Related work}
\label{sec:related}
\vspace{-1mm}

We discuss previous work that is most relevant to ours in the fields of 
texture transfer and synthesis for 3D shapes.

\vspace{-5mm}
\paragraph{Template-based methods} assume that all target shapes can be represented by deforming a template mesh, usually a sphere~\cite{pavllo2020convolutional,bhattad2021view,henderson20cvpr,tulsiani2020implicit,vmr2020,dibr} or a plane~\cite{pan2020gan2shape,wu2020unsupervised}. The UV mapping of the template mesh is given and transferred to the target shape after deformation.
However, by imposing a mesh template, these methods often cannot capture details, especially when the topology or the structure of the target shape is complex. 
For example, when deforming a sphere into a human body, it is hard to accurately reconstruct the fingers. Even if the deformation is successful, the texture of the fingers, when projected from the human body to a sphere and then to its texture image, is typically heavily distorted.

\vspace{-5mm}
\paragraph{UV map from artists.}
Another line of work~\cite{yin2021_3DStyleNet,chaudhuri2021semi} does not assume template meshes are given, but instead assumes that the UV maps are provided with the 3D shapes. The UV maps and textures are typically modeled by artists and can be in arbitrary layouts. To address this issue, these methods usually require ground-truth semantic segmentation of the texture image or the 3D shape for semantic-aware texture synthesis. In our work, we aim to perform texture synthesis without such supervision.
Automatic UV mapping has also been extensively studied in computer graphics, though for single shapes. It includes mesh parameterization with certain constraints, and surface cutting to generate charts with disk topology. We refer to \cite{sheffer2007mesh} for a survey of related techniques.
Different from these traditional methods, we learn aligned UV maps for a set of shapes.

\vspace{-5mm}
\paragraph{Discretization and colorization.}
Instead of adopting UV maps to reduce the dimensionality of textures from 3D to 2D, some methods discretize the 3D shapes into ``atoms'' and then colorize each ``atom''. When a shape is represented as a voxel grid, the shape can be textured by predicting the color of each voxel~\cite{chen2018text2shape,sun2018im2avatar}.  For triangle meshes, the color of each vertex can be predicted~\cite{gao2020learning}. However, since  discretization is in 3D rather than in 2D (pixels), these approaches either cannot scale up, or cannot predict the color efficiently due to the irregularity of the representation.

\vspace{-5mm}
\paragraph{Texture fields}~\cite{texture_fields} predict the color for each 3D point in a continuous 3D space. The NeRF family~\cite{mildenhall2020nerf} also adopts this approach, by using the viewing direction as an additional condition for predicting the color of each point. Since the NeRF family does not directly generate textures for 3D shapes, we mainly discuss and compare with Texture Fields in this paper. One major issue of Texture Fields is that it is unable to represent high-frequency details, which is a property of the MLPs that it uses. Positional encoding ~\cite{mildenhall2020nerf} and SIREN ~\cite{sitzmann2020implicit} are proposed to alleviate this issue, which works well on overfitting of single shapes. However, performance degrades significantly in generative tasks. The results of implicit methods tend to be smooth and lack high-frequency details~\cite{sitzmann2020implicit,chan2021pi}.

\vspace{-5mm}
\paragraph{Shape correspondences.} There is a large body of work that finds dense correspondences among shapes~\cite{liu2020learning,deng2021deformed}, which  can also enable texture transfer. However, these methods do not take color into account when finding correspondences, 
which may hinder their performance. 

\vspace{-1.5mm}
\section{Our Approach}
\label{sec:method}
\vspace{-1mm}

\begin{figure*}[t!]
\begin{center}
\includegraphics[width=1.0\linewidth]{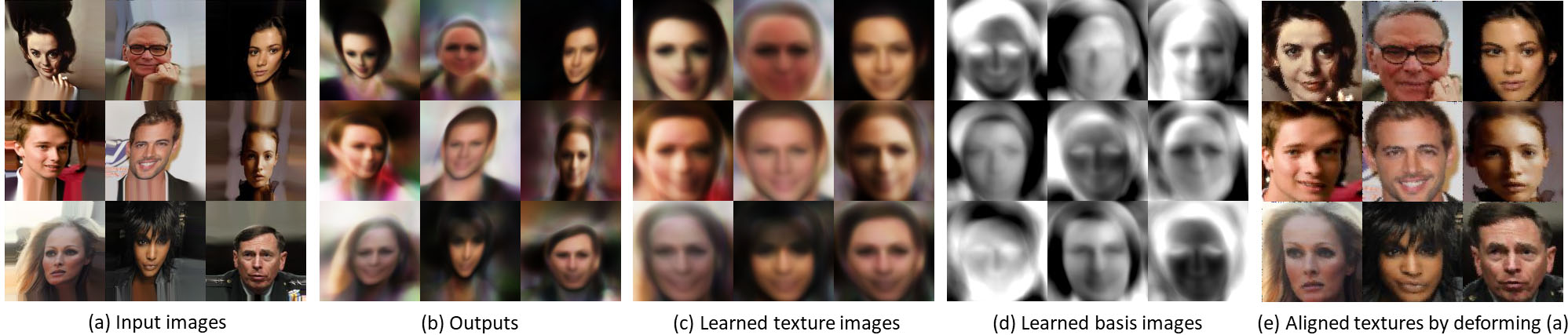}
\end{center}
\caption{
Results of the 2D toy experiment on the face dataset. Our network reconstructs input images (a) by learning a set of basis images (d) and linearly combining them into aligned texture images (c), and then deforming the texture images (c) into the outputs (b) via learned UV mapping. The learned UV mapping can be used to deform the input images (a) into aligned high-quality texture images (e).}
\label{fig:toy_result}
\end{figure*}
\begin{figure}[t!]
\begin{center}
\includegraphics[width=1.0\linewidth]{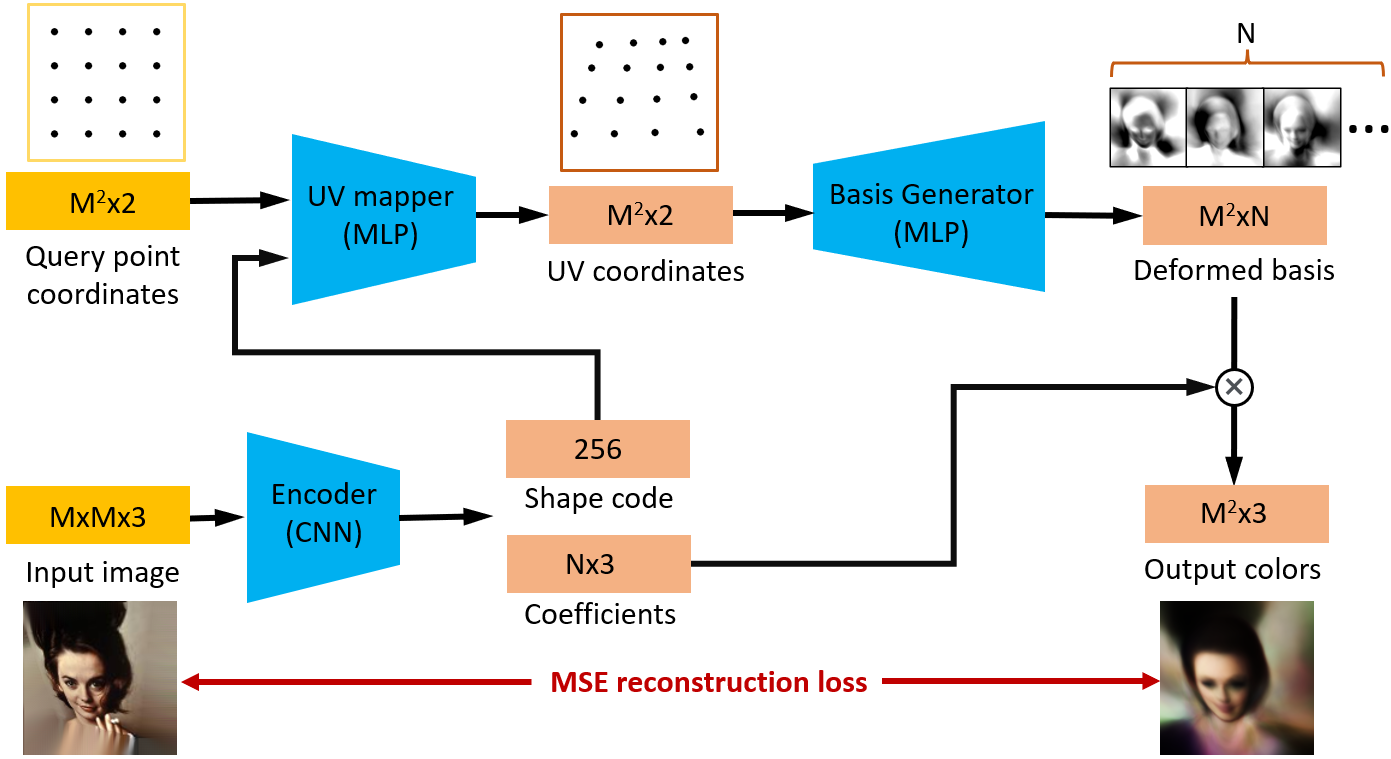}
\end{center}
\caption{
Network architecture of the 2D toy experiment on the face dataset, to demonstrate the concept of our alignment module.}
\label{fig:toy_net}
\end{figure}

In this section, we first introduce our core alignment module in 2D and verify it with a 2D-to-2D image alignment experiment in Sec~\ref{sec:method_2d}.
We then explain our network architecture for learning aligned UV maps for 3D shapes in Sec~\ref{sec:method_3d}.
Finally, in Sec~\ref{sec:exps_app} we show applications enabled by our approach, including texture transfer, texture synthesis, and texture prediction from single images.

\vspace{-1mm}
\subsection{Texture alignment module}
\label{sec:method_2d}
\vspace{-1mm}

Learning aligned textures for a set of shapes is a complex task. However, we show that a simple alignment module performs surprisingly well.
Before we introduce our network for 3D shapes, we will use a toy 2D experiment to demonstrate how the alignment module works.

\vspace{-5mm}
\paragraph{Task.}
Given a set of face images in random poses, as shown in Fig.~\ref{fig:toy_result}(a), we aim to align them into a canonical pose, as shown in Fig.~\ref{fig:toy_result}(e). We take 1,000 face images from CelebA-HQ ~\cite{karras2017progressive,liu2015faceattributes} and perform random perspective transformations to obtain 128x128 training images, such as those in Fig.~\ref{fig:toy_result}(a).
To link this task with texture mapping, one could consider training images in Fig.~\ref{fig:toy_result}(a) as square shapes in 2D, and Fig.~\ref{fig:toy_result}(e) are their aligned texture images. The color of each pixel in the square shape must be retrieved from the shape's texture image, by mapping the pixel's coordinates into the UV space to get the UV coordinates, and then indexing the texture image with those UV coordinates.

\vspace{-5mm}
\paragraph{Insight.}
We take inspiration from classic linear subspace learning methods such as eigenfaces~\cite{Turk1991}, where a basis is computed via PCA for a set of face images, so that each face is decomposed into a weighted sum of the eigenfaces. Note that PCA works best when the images are aligned. Therefore, if a network is designed to decompose the input images into weighted sums of basis images, and is allowed to deform the input images before the decomposition, the network should learn to align the input images into a canonical pose, and decompose the aligned images so as to minimize the reconstruction error.

\begin{figure*}[t!]
\begin{center}
\includegraphics[width=1.0\linewidth]{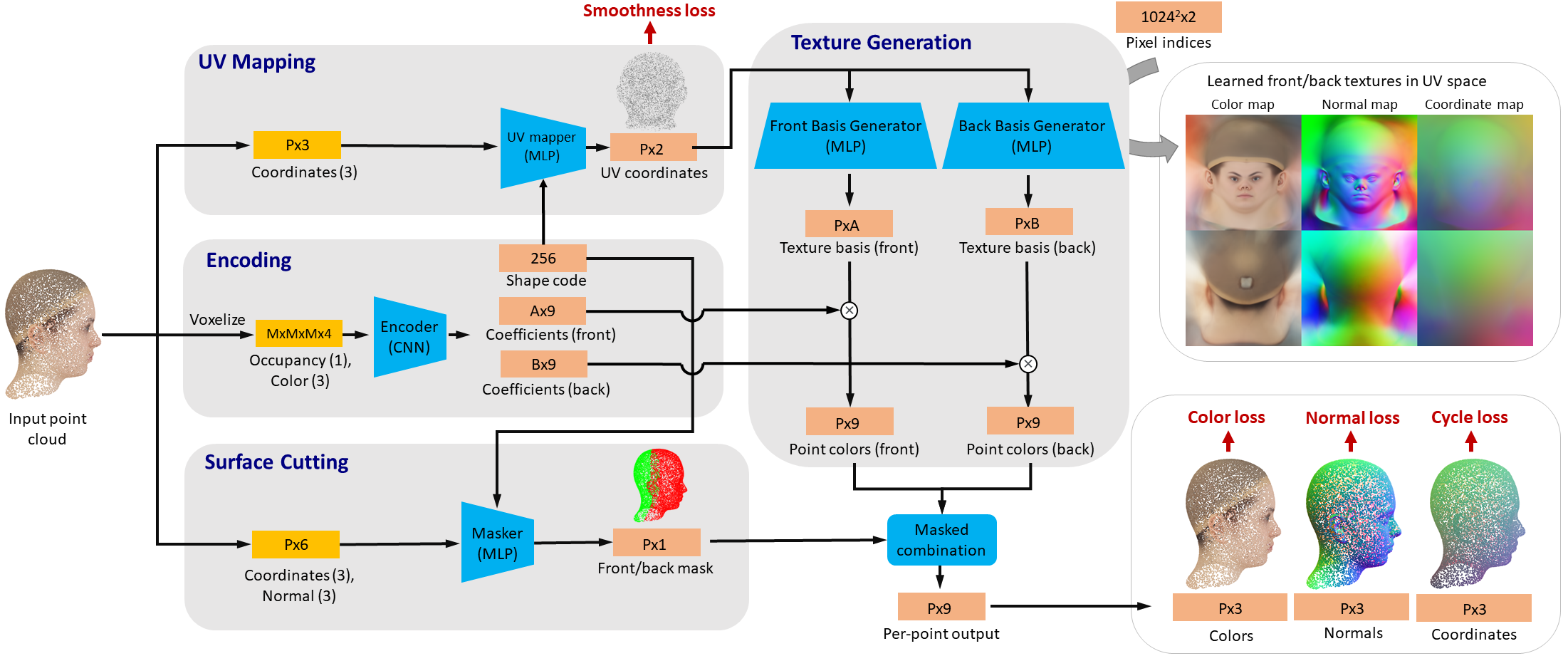}
\end{center}
\caption{
Network architecture of our AUV-Net.
The encoder predicts the shape code and the coefficients from the voxelized input point cloud. The UV mapper and the masker take as input the shape code and the query points from the input point cloud, and output the UV coordinates and the segmentation mask, respectively. The UV coordinates are fed into the two basis generators to obtain the basis colors for each query point, and the basis colors are multiplied by the predicted coefficients to generate the actual colors for each query point. Those colors from the two basis generators are selected by the predicted segmentation mask to produce the final colors.
}
\label{fig:network}
\end{figure*}

\vspace{-5mm}
\paragraph{Framework.}
Fig.~\ref{fig:toy_net} illustrates our alignment module operating for 2D images.
It is composed of three neural networks: a \emph{basis generator} to predict a set of basis images; an \emph{encoder} to predict the coefficients to weigh the basis images; and a \emph{UV mapper} to predict the UV coordinates for each query point. The encoder also predicts a shape code to condition the UV mapper.

\vspace{-5mm}
\paragraph{Basis generator.}
Our basis generator is a Multilayer Perceptron (MLP) that takes a 2D point $(x,y)$ as input and outputs the color of this point. For $N$ basis images, the network outputs $N$ values ($N$ gray-scale colors). In this toy 2D experiment, we use $N=128$. We adopt an MLP for generating the basis because it is fully differentiable with respect to both the colors of the basis and the input point coordinates.
In contrast, if a Convolutional Neural Network (CNN) or a grid of learnable weights is applied to generate the basis, one would need to index the output grid with query points, limiting the gradients from the output color to the basis and the query point coordinates to small neighborhoods.

\vspace{-5mm}
\paragraph{UV mapper.}
Once trained, we can input a regular grid of query points to the basis generator to obtain the aligned basis images such as those in Fig.~\ref{fig:toy_result}(d), 
and the aligned texture images shown in Fig.~\ref{fig:toy_result}(c) by multiplying the basis images with the coefficients.
However, at training time, we need to ``deform'' the texture images to reconstruct the input images in Fig.~\ref{fig:toy_result}(a). This is achieved by the UV mapper, which maps the query points sampled from the square shape into UV coordinates to index the texture image, as shown in Fig.~\ref{fig:toy_net}. The final deformed outputs are shown in Fig.~\ref{fig:toy_result}(b).
The UV mapper is an MLP conditioned on a shape latent code. It takes 2D point coordinates concatenated with the shape code as input, and outputs UV coordinates.

\vspace{-5mm}
\paragraph{Encoder and loss function.}
Since the inputs are images, we use a 2D CNN as our encoder to predict shape codes and coefficients. 
The deformed basis produced by the basis generator is multiplied with the coefficients to produce the final output, as shown in Fig~\ref{fig:toy_net}.
We use Mean Squared Error (MSE) between the output and the input image as the reconstruction loss.
During early stage of training (first few epochs), we apply a prior loss, i.e., MSE between the query point coordinates and their corresponding UV coordinates, to encourage the UV mapper to perform an identity mapping, so that the basis is initialized with appropriate orientation, scale, and position.

\vspace{-5mm}
\paragraph{Obtaining high-quality texture images.}
After training, our network produces aligned texture images as shown in Fig.~\ref{fig:toy_result}(c). However, the images are in low quality since they are constructed by a limited number of over-smoothed basis images.
To obtain a high-quality texture image with details, we can deform the input image into the UV space. We sample points from the input image, feed those points to the UV mapper to obtain UV coordinates, use the UV coordinates and colors of the sampled points to fill a blank image, and finally inpaint the missing regions. We use~\cite{telea2004image} for inpainting. The results are  shown in Fig.~\ref{fig:toy_result}(e).

\subsection{Learning aligned UV maps for 3D shapes}
\label{sec:method_3d}
\vspace{-1mm}

Our network for 3D shapes, dubbed AUV-Net, is built upon the alignment module in Fig.~\ref{fig:toy_net}. It can be thought of as a 3D-to-2D version of the 2D-to-2D alignment module, with modifications to address issues caused by the properties of 3D shapes. The architecture of AUV-Net is shown in Fig.~\ref{fig:network}. In the following, we will first describe the notable changes in AUV-Net compared to the 2D alignment module, and then the loss functions and training details.

\vspace{-5mm}
\paragraph{Predicting color, normal, and coordinate maps.}
3D shapes tend to have textures with large areas of pure color or similar colors. Those featureless regions make it hard for our network to align the textures with only a loss function defined on colors. Therefore, in addition to the color maps, our network also produces normal maps and coordinate maps, as shown in Fig.~\ref{fig:network} top-right.
The normal maps are used for predicting the unit normals of input points. The coordinate maps are used for predicting the positions of the input 3D points,
therefore forming a 3D-2D-3D cycle in our network, which encourages injective UV mapping.

\vspace{-5mm}
\paragraph{Cutting surfaces with a masking network.}
Unlike images, it is usually impossible to embed a 3D shape onto a 2D plane without overlap or severe distortion.
Therefore, we introduce a \emph{masker}, to generate a segmentation mask for the input shape, as shown in Fig.~\ref{fig:network} bottom-left. This can be considered as cutting the 3D shape into multiple pieces, so that each piece can be represented by a single texture image. The input to the masker contains point coordinates, point normals, and the shape code. The normals are essential to segmenting the shapes, since thin parts such as fingers on a human body mesh are very hard to segment with only point coordinates. The predicted segmentation mask ($M$) is used to mask the outputs of the two basis generators ($A$ and $B$), as $M \cdot A + (1-M) \cdot B$, to produce the final output.

\vspace{-5mm}
\paragraph{Multiple basis generators.}
We introduce two basis generators to represent the ``front'' and the ``back'' part of a shape, respectively. The ``front'' does not have to literally denote the front-facing part of a shape. It simply refers to a part of the shape so that the union of the ``front'' and ``back'' covers the entire shape. The outputs of the two basis generators are shown in Fig.~\ref{fig:network} right, where the head is being represented with two texture maps. Note that the number of basis generators does not have to be two; we use four basis generators for chairs in our experiments.

\vspace{-5mm}
\paragraph{Shared UV mapper.}
We use one shared UV mapper for both the front and the back basis generators, instead of two independent UV mappers. This is based on a careful consideration. In our experiments, one of the most prominent issues when we transfer the texture from one shape to another is that we inevitably obtain seams between the two pieces of shapes using two different texture images. A shared UV mapper alleviates this issue by forcing the two pieces to share the same boundary in the texture images.
It does not fully resolve the seam issue, but it is very helpful in practice. After we inpaint the texture images, the seams are barely visible in most cases.

\vspace{-5mm}
\paragraph{Loss functions.} 
To train AUV-Net, the meshes are converted into point clouds with normals and colors, as input to our network. We also voxelize the point clouds to obtain colored voxel grids as input to the 3D CNN encoder.
The overall loss function is composed of five terms:\\[-2mm]
\begin{equation}
L = w_{c}L_{c} + w_{n}L_{n} + w_{x}L_{x} + w_{s}L_{s} + w_{p}L_{p}
\label{eq:allterms}
\end{equation}
where $L_{c}$, $L_{n}$, $L_{x}$ denote the color loss, the normal loss, and the cycle consistency loss on the 3D coordinates, respectively. They are defined as MSEs between the predictions and the ground truth.

$L_{s}$ is the smoothness loss. For a subset of input points, we find their neighbors within a distance $\sigma=0.02$ , and use the distances between the points and their neighbors to regularize the corresponding distances in the UV space:\\[-2mm]
\begin{equation}
L_{s} = \frac{1}{MN} \sum_{i=1}^{M} \sum_{j=1}^{N} | D(p_i,p_j) - D(q_i,q_j) | \cdot T(p_i,p_j)
\end{equation}
where $N$ is the number of input points, $M$ is the size of the subset, $p_i$ is the $i$-th input 3D point, 
$q_i$ is the 2D UV point predicted for $p_i$ by the UV mapper.
$D(a,b)$ is the Euclidean distance between point $a$ and $b$. $T(a,b)$ is defined as $1$ if $D(a,b)<\sigma$, and $0$ otherwise. In each mini-batch, we process one shape, with $N = 16,384$ and $M = 2,048$.

$L_{p}$ is the prior loss to initialize the UV coordinates and the masks. It may vary per category of the training shapes. For the human head dataset shown in Fig.~\ref{fig:teaser} and Fig.~\ref{fig:network}, where all the heads are facing z direction, we have:\\[-2mm]
\begin{equation}
L_{p} = \frac{1}{N} \sum_{i=1}^{N} (p_i^x - q_i^x)^2 + (p_i^y - q_i^y)^2 + (m_i - n_i)^2
\end{equation}
where $p_i^x$ is the x coordinate of $p_i$, $q_i^x$ is the x coordinate of $q_i$, $m_i$ is the masking value predicted by the masker for $p_i$. $n_i$ is defined as $1$ if the unit normal of $p_i$ in the z direction is greater than $-0.5$, and $0$ otherwise. 
This prior loss initializes the UV mapping by projecting the 3D points onto the xy-plane.
To cut the shape into two pieces, this prior loss follows our prior that: if the angle between a point's normal and z axis is less than $120$ degrees, the point belongs to the ``front'' part. Similar to Sec.~\ref{sec:method_2d}, prior loss is only used in the first few epochs of training to initialize the mask and the UV coordinates. We provide the definitions of prior losses for other categories in the supplementary.

\vspace{-5mm}
\paragraph{Assumptions on the training set.} 
Note that the above loss terms assume certain properties of the training dataset.
First, the shapes need to have part-level correspondences, as the network actually assigns dense correspondences between shapes when it maps all shapes into the same UV space. Therefore, we only train our model on 3D shapes of the same category. Second, the shapes need to be pose aligned, e.g., heads should all face z direction in the aforementioned human head dataset. We also normalize all shapes to unit boxes before training, to avoid interference of drastically different scales.

\vspace{-5mm}
\paragraph{Multi-stage training.} 
We train the network in three stages, due to a trade-off between the quality of the texture alignment and the level of distortion. In some cases, aligning textures requires heavy distortion in the texture images, e.g., when aligning a sedan with a van (Fig.~\ref{fig:transfer}). However, less distortion is a desirable feature that reduces aliasing effect when rendering the textures, and makes post-processing easier, e.g., when being edited by an artist. We find that if the network is trained with fixed weighting of the loss terms, we cannot get both the alignment and minimal distortion. 
Therefore, we first initialize the network with prior loss $L_{p}$ and a set of weights focused on minimal distortion. In the second stage, we remove $L_{p}$, and use weights focused on alignment. In the final stage, we use weights focused on minimal distortion. For the human head dataset, the first stage has 10 epochs, with $\{ w_{c},w_{n},w_{x},w_{s},w_{p} \} = \{ 1,0.5,100,100,1 \}$;  second stage has 2,000 epochs, with $\{ 1,0.5,1,1,0 \}$; third stage has 2,000 epochs, with $\{ 1,0.5,100,100,0 \}$. Training takes 2 days on one NVIDIA RTX 3080 Ti GPU. Other training details are in the supplementary.

\subsection{Applications}
\label{sec:exps_app}
\vspace{-1mm}

\paragraph{Texture transfer.} 
After training AUV-Net, we obtain aligned high-quality texture images ($1024^2$ in our experiments) for all training shapes, as shown in Fig.~\ref{fig:teaser}. 
The fact that these texture images are aligned allows us to transfer textures between two training shapes by simply swapping their texture images, as shown in Fig.~\ref{fig:teaser} and ~\ref{fig:transfer}.
We denote this application as {\bf Tsf (transfer)}. 
Given a new shape that is not in the training set, we can also texture it by mapping its vertices into the aligned UV space. This is done via a post-training optimization stage, in which we add the new shape into the training set, and continue training the network for a few epochs. 
During the optimization, we fix the weights of the basis generators to reuse the well-learned texture basis.

\vspace{-5mm}
\paragraph{Texture synthesis.} 
A great advantage of having aligned texture images is that it allows us to utilize existing 2D generative models to synthesize new textures for 3D shapes. 
We train StyleGAN2 ~\cite{Karras2019stylegan2} in experiments and show results in Fig.~\ref{fig:gen}. We denote this application as {\bf Gen (generation)}.

\vspace{-5mm}
\paragraph{Single-image 3D reconstruction.} We can condition texture synthesis on a variety of inputs, for example, reconstructing textured 3D shapes from single images, as shown in Fig.~\ref{fig:svr}. To this end, we add a 2D ResNet~\cite{resnet} image encoder to predict the texture latent code and the shape code from an input image, a CNN decoder to predict the aligned texture images from the texture latent code, and an IM-Net decoder~\cite{imnet} to predict the geometry of the shape conditioned on the shape code. We denote this application as {\bf SVR (single view reconstruction)}. Implementation details are provided in the supplementary.
\vspace{-1mm}
\section{Experiments}
\label{sec:exps}

\vspace{-1mm}
\paragraph{Datasets.} 
We use six datasets in our experiments, as listed in Table~\ref{table:data_stats}. Information about  dataset licenses is in the supplementary. 
We mainly perform generative tasks (Gen, SVR) on ShapeNet~\cite{chang2015shapenet} categories since other datasets have too few training shapes. 
Note that the original shapes in ShapeNet usually have complex geometry but simple textures. We create a new version of ShapeNet Cars and Chairs better suited for the texture transfer/synthesis task, by simplifying the meshes to reduce geometric details and baking the geometric details into textures.

\vspace{-1mm}
\subsection{Texture Transfer}
\label{sec:exps_Tsf}
\vspace{-1mm}

\begin{figure}[t!]
\begin{center}
\includegraphics[width=1.0\linewidth]{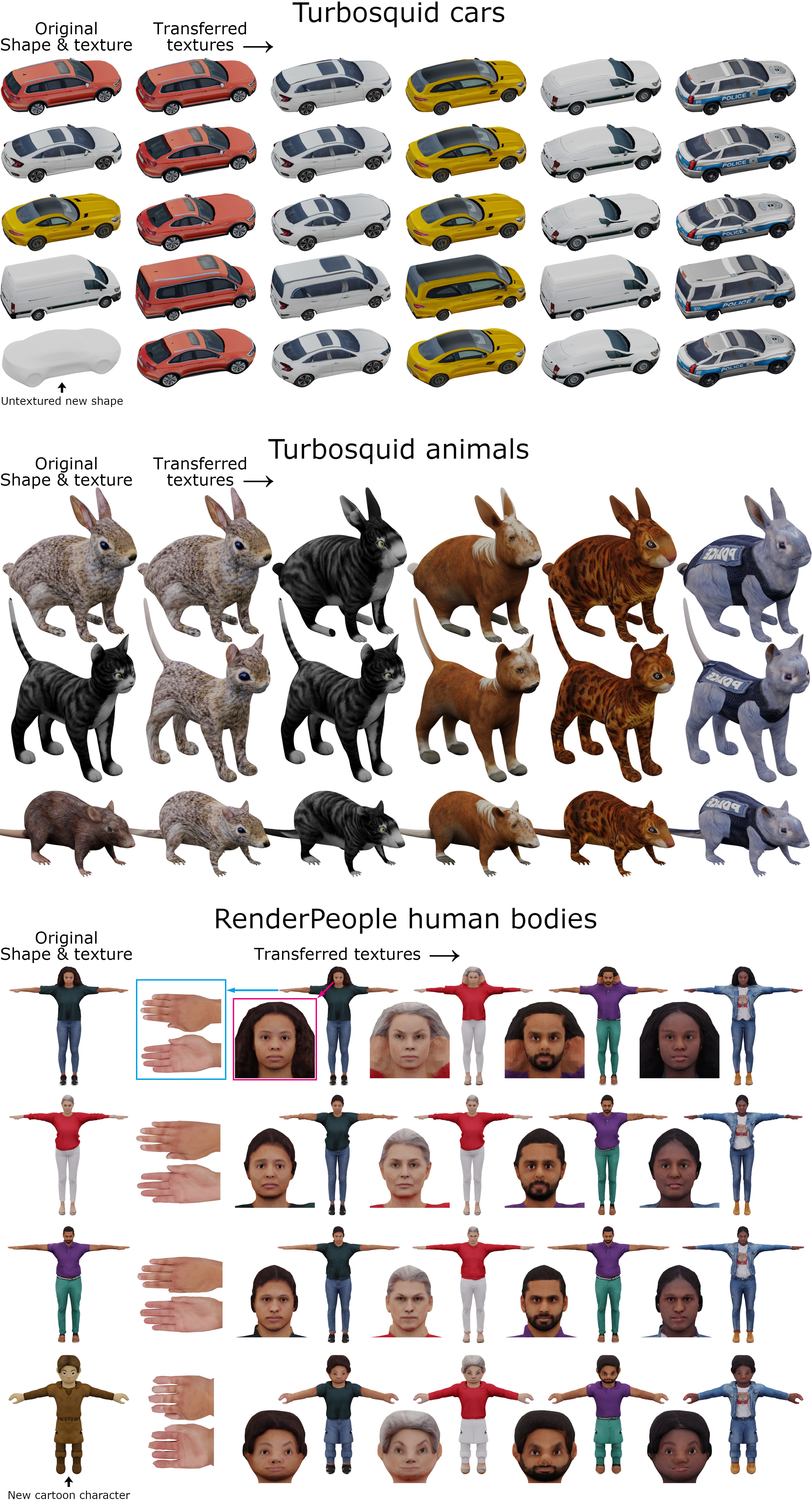}
\end{center}
\caption{
Texture transfer results. We show three categories in this figure: Turbosquid cars (top), Turbosquid animals (middle), and RenderPeople human bodies (bottom). Triplegangers heads can be found in Figure~\ref{fig:teaser}. For RenderPeople, we show a zoom-in of the head on the lower left of each shape; we also show zoom-ins of hands for the second column of shapes.
}
\label{fig:transfer}
\end{figure}

\begin{figure}[t!]
\begin{center}
\includegraphics[width=1.0\linewidth]{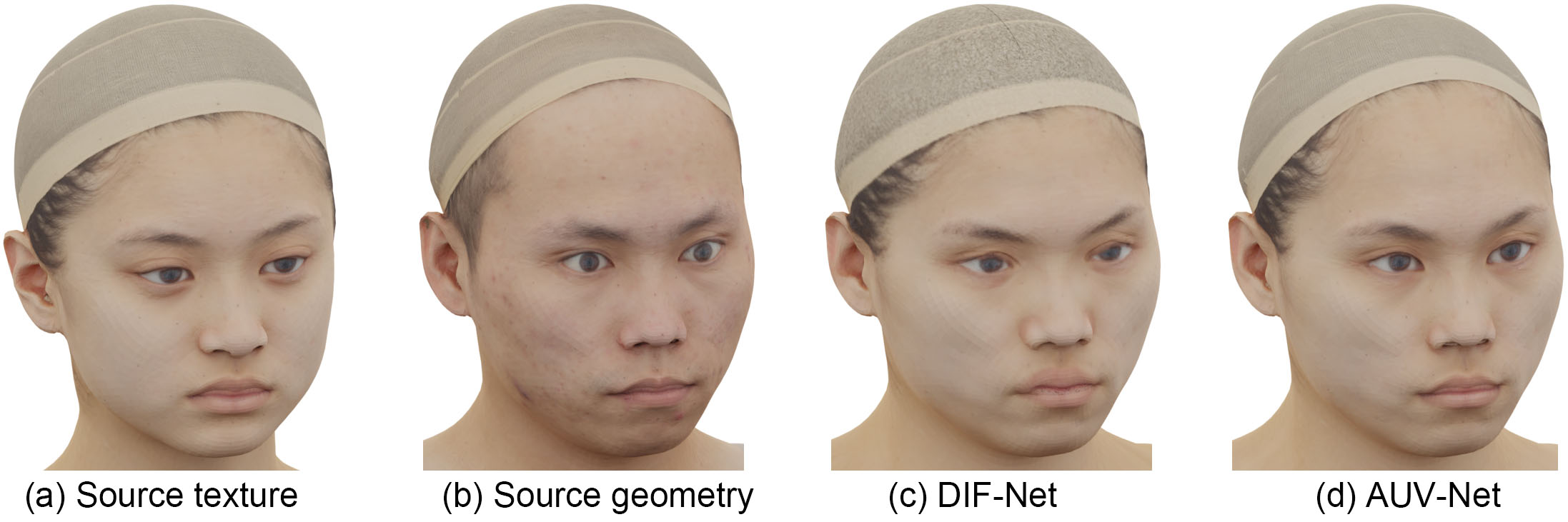}
\end{center}
\caption{
Comparison with DIF-Net~\cite{deng2021deformed} on texture transfer. The texture is transferred from (a) to (b). In (c), the eyes' shapes are not changed with respect to (a), the lips are misaligned, and the hat is lower than it should be compared to (b). Those details are mostly represented in colors rather than geometry.
}
\label{fig:compareDIF}
\end{figure}

We show texture transfer results in Fig.~\ref{fig:teaser} and ~\ref{fig:transfer}. Only non-ShapeNet categories are shown due to page limit. More results can be found in the supplementary. Our method uses cues such as colors, normals, and positions when learning aligned UV mapping, 
and therefore performs well on aligning facial orifices, car windows and wheels, fingers, and animal limbs.
Previous methods find dense correspondences among shapes by deforming geometry ~\cite{liu2020learning,deng2021deformed}. However, they do not utilize color information, and may thus misalign regions with fewer geometric cues, as shown in Fig.~\ref{fig:compareDIF}.
We show results on transferring textures to new shapes that are clearly different from the training shapes in Fig.~\ref{fig:transfer}. Our method is able to correctly texture an over-simplified texture-less car model, and transfer textures to a cartoon character model.

\vspace{-3mm}

\begin{table}[!t]
\begin{center}
\resizebox{1.0\linewidth}{!}{
\begin{tabular}{l|c|c}
\hline
Dataset name & Number of Shapes & Applications \\
\hline
ShapeNet ~\cite{chang2015shapenet} cars & 7,497 & Tsf, Gen, SVR  \\
ShapeNet ~\cite{chang2015shapenet} chairs & 6,778 & Tsf, Gen, SVR \\
Turbosquid ~\cite{Turbosquid} cars & 436 & Tsf \\
RenderPeople ~\cite{RenderPeople} human bodies & 500 & Tsf \\
Triplegangers ~\cite{Triplegangers} heads & 515 & Tsf, Gen \\
Turbosquid ~\cite{Turbosquid} animals & 442 & Tsf \\
\hline
\end{tabular}
}
\end{center}
\caption{Datasets used in our experiments. Tsf, Gen, and SVR refer to the applications listed in Sec.~\ref{sec:exps_app}. }
\label{table:data_stats}
\end{table}

\begin{figure}[t!]
\begin{center}
\includegraphics[width=1.0\linewidth]{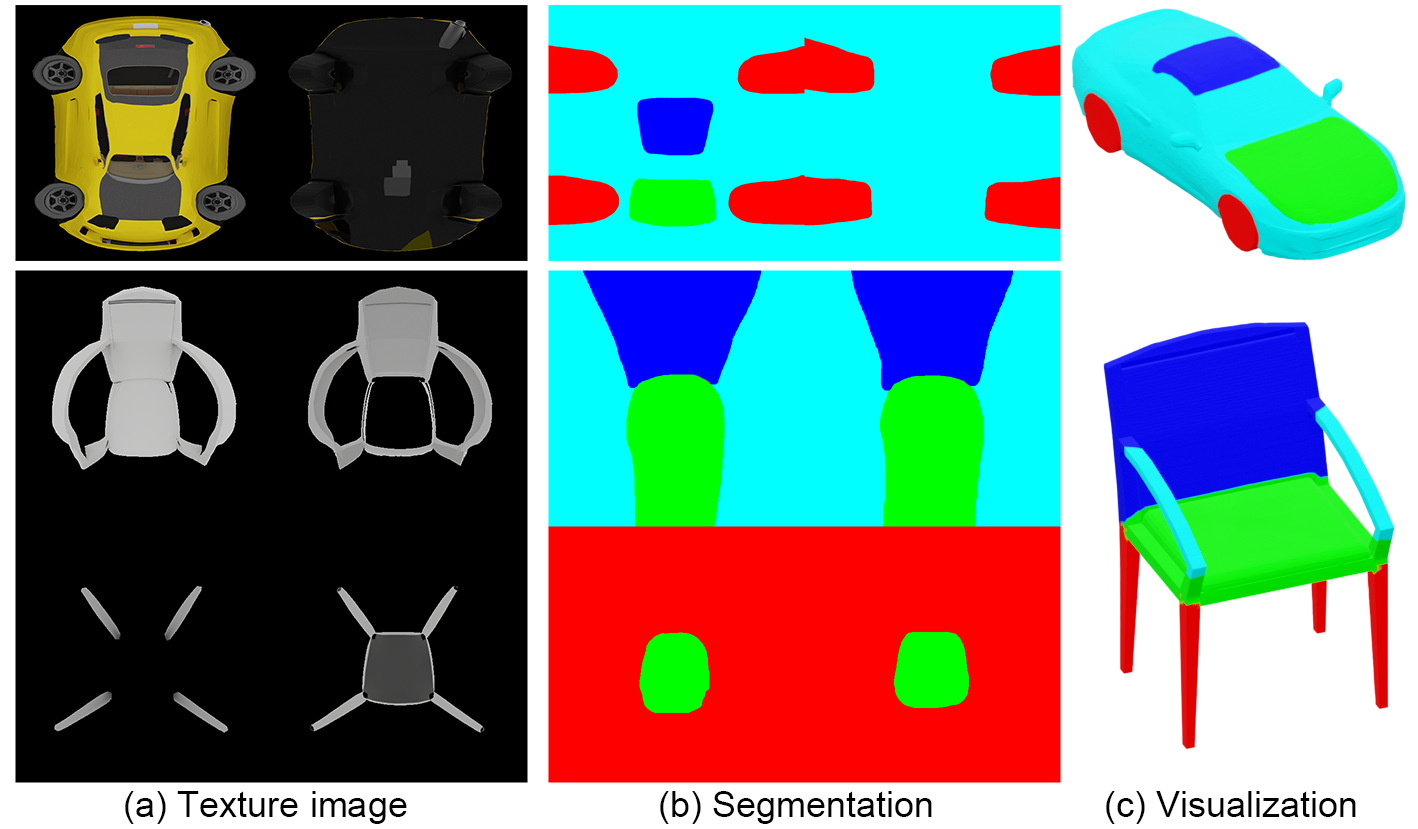}
\end{center}
\caption{
Sample texture images and segmentation on ShapeNet cars and chairs. (a) shows texture images before inpainting. Note that there are 2 texture images for each car and 4 for each chair. In (b), we show the segmentation we used to produce Table~\ref{table:tsf_numbers}. A visualization on 3D shapes is shown in (c).
}
\label{fig:seg}
\end{figure}

\paragraph{Quantitative evaluation.} To evaluate the alignment quality, we label one texture image with a different color per semantic part, as shown in Fig.~\ref{fig:seg} (b). Since the texture image is aligned across shapes, we get semantic segmentation of 3D shapes with a single labeled example. We evaluate our part segmentation of shapes with ground truth segmentation provided in the ShapeNet part dataset~\cite{yi2016scalable}. We compare with BAE-Net~\cite{chen2019bae_net} that performs one-shot shape segmentation and DIF-Net~\cite{deng2021deformed} that learns dense correspondences, and report Intersection Over Union (IOU) in Table~\ref{table:tsf_numbers}. Our method outperforms alternatives.

\begin{table}[!t]
\begin{center}
\resizebox{1.0\linewidth}{!}{
\begin{tabular}{l|c|c}
\hline
Dataset (\#parts) & ShapeNet cars (4) & ShapeNet chairs (4) \\
\hline
Segmented parts & Wheel, body, hood, roof & Back, seat, leg, arm \\
\hline
BAE-Net & 59.3  & 85.2   \\
DIF-Net  & 69.0  & 80.3 \\
AUV-Net     & {\bf 72.7}  & {\bf 85.8} \\
\hline
\end{tabular}
}
\end{center}
\caption{Semantic segmentation results in IOU, comparing with BAE-Net~\cite{chen2019bae_net} and DIF-Net~\cite{deng2021deformed}. }
\label{table:tsf_numbers}
\end{table}

\begin{figure}[t!]
\begin{center}
\includegraphics[width=1.0\linewidth]{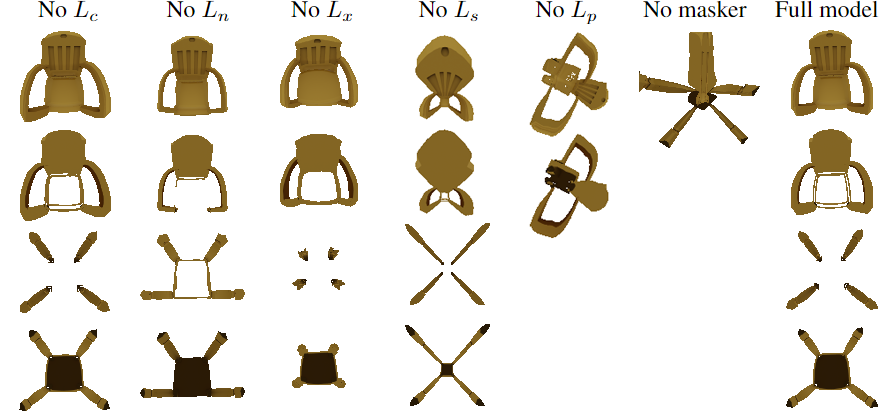}
\end{center}
\caption{
Ablation study: learned texture images with different settings. Each chair has four texture images (shown vertically).
}
\label{fig:ablation_chair}
\end{figure}

\vspace{-5mm}
\paragraph{Ablation study.} We provide the ablation study in Table~\ref{table:ablation_numbers} and Fig.~\ref{fig:ablation_chair}, where we remove one of the five loss terms in Eq.~\ref{eq:allterms}, or the masker module. We use the same evaluation setting described above. The results are consistent with our design choices of the individual modules. The color loss $L_c$, the normal loss $L_n$, and the cycle loss $L_x$ are designed to help find correspondences, therefore removing them often causes the performance to drop (Table~\ref{table:ablation_numbers}). The smoothness loss $L_s$ is designed to regularize the UV coordinates; it may hurt the correspondence, but removing it can cause certain parts to be squished (Fig.~\ref{fig:ablation_chair} column 4), thus making texture synthesis difficult. The cycle loss $L_x$ also helps regularize the UV coordinates since it encourages one-to-one mapping between surfaces of 3D shapes and the 2D textures; removing it causes overlap in the texture images (Fig.~\ref{fig:ablation_chair} column 3). The prior loss $L_p$ and the masker are critical to our model, as removing them causes severe segmentation and overlap issues (Fig.~\ref{fig:ablation_chair} column 5\&6). Also note that different categories have different sensitivity to the loss terms. As shown in Table~\ref{table:ablation_numbers}, the car category relies heavily on the colors: removing $L_c$ leads to significant performance drop. In contrast, chair category is less sensitive: removing $L_c$ has no visible effect on the learned texture maps in Fig.~\ref{fig:ablation_chair}.

\begin{table}[!t]
\begin{center}
\resizebox{1.0\linewidth}{!}{
\begin{tabular}{lccccccc}
\hline
 & No $L_c$ & No $L_n$ & No $L_x$ & No $L_s$ & No $L_p$ & No masker & Full model \\
\hline
Cars & 68.5 & 71.7 & {\bf 73.0} & 72.8 & 70.6 & 72.0 & 72.7 \\
Chairs & 85.2 & 84.6 & 83.7 & {\bf 87.1} & 85.7 & 71.1 & 85.8 \\
\hline
\end{tabular}
}
\end{center}
\caption{Ablation study: semantic segmentation results in IOU. }
\label{table:ablation_numbers}
\end{table}

\vspace{-1mm}
\subsection{Texture Synthesis}
\label{sec:exps_Gen}
\vspace{-1mm}

\begin{figure*}[t!]
\begin{center}
\includegraphics[width=1.0\linewidth]{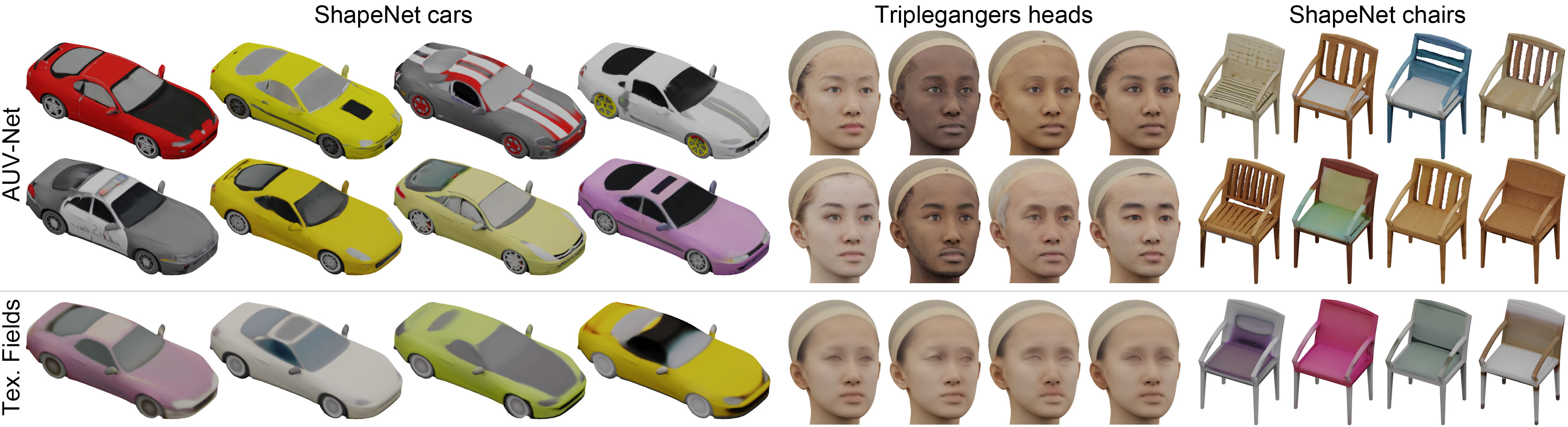}
\end{center}
\caption{
Texture synthesis results. The holes on chairs are hallucinated via texture transparency (alpha channels in the texture images).
}
\label{fig:gen}
\end{figure*}

\begin{table}[!t]
\begin{center}
\resizebox{1.0\linewidth}{!}{
\begin{tabular}{l|c|c|c}
\hline
 & Triplegangers & ShapeNet cars & ShapeNet chairs \\
\hline
Tex. Fields & 24.59  & 53.09  & 7.03  \\
AUV-Net        &  {\bf 5.69}   &  {\bf 12.11}  &  {\bf 5.33}  \\
\hline
\end{tabular}
}
\end{center}
\caption{Quantitative results of generative models in FID. }
\label{table:gen_numbers}
\end{table}

We show texture synthesis results in Fig.~\ref{fig:gen}, and compare them with Texture Fields~\cite{texture_fields} (TF). 
Our method generates more details and gets correct alignments, while TF outputs smooth color chunks that are sometimes misaligned. This is because TF uses an MLP to map continuous 3D coordinates into colors, and does not properly disentangle texture and geometry.
In contrast, Our method uses a sophisticated 2D generative model trained on aligned texture images to generate the outputs. The texture images aligned by AUV-NET are mostly independent from the actual mesh geometry.

To evaluate the methods quantitatively, we use Fr\'echet Inception Distance (FID)~\cite{heusel2017gans}. We test on 1,000 shapes for each ShapeNet category and 100 shapes for Triplegangers heads. For each test shape, we generate 5 textures, and render each textured shape into 8 views. Results are presented in Table~\ref{table:gen_numbers}, where our method clearly outperforms baseline.

\vspace{-1mm}
\subsection{Textured Single-View 3D Reconstruction}
\label{sec:exps_SVR}
\vspace{-1mm}

\begin{figure}[t!]
\begin{center}
\includegraphics[width=1.0\linewidth]{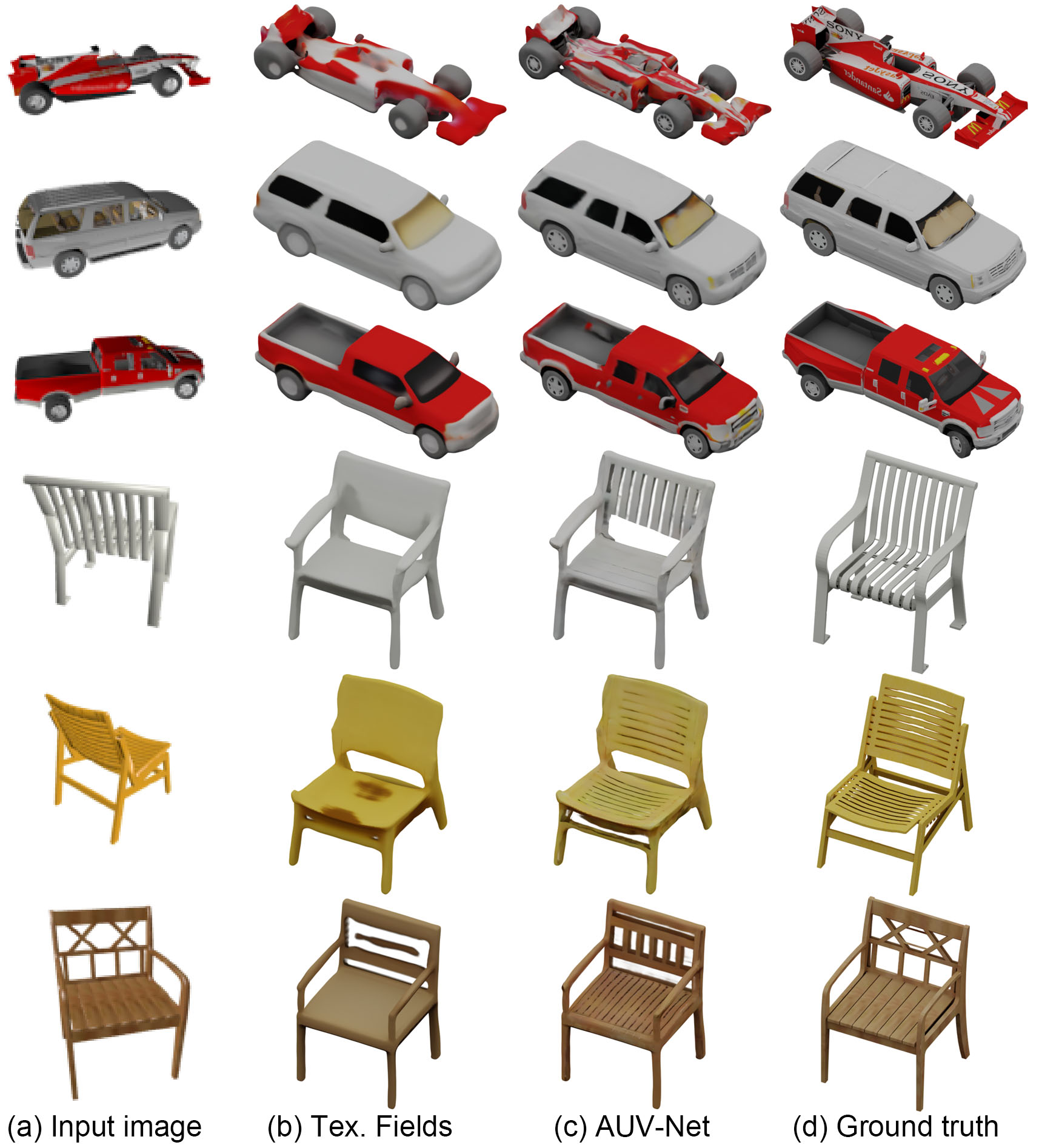}
\end{center}
\caption{
Textured single view reconstruction results. Zoom in to see the details, e.g., wheels of the cars.
}
\label{fig:svr}
\end{figure}

We show results in Fig.~\ref{fig:svr}, and compare with TF. For a fair comparison on texture prediction, we use the mesh generated by our method as the output mesh of TF, so that TF only needs to predict textures of the shapes. The results show that our method produces sharper boundaries and more details in the textures.

\begin{table}[!t]
\begin{center}
\resizebox{1.0\linewidth}{!}{
\begin{tabular}{l|c|c|c|c|c|c}
\hline
 & \multicolumn{3}{|c|}{ShapeNet cars} & \multicolumn{3}{c}{ShapeNet chairs} \\
\hline
 & FID & SSIM & feature-$l_1$ & FID & SSIM & feature-$l_1$ \\
\hline
Tex. Fields & 92.89 & {\bf 0.897} & 0.219 & 36.89 & {\bf 0.855} & 0.193   \\
AUV-Net        & {\bf 40.85} & 0.894 & {\bf 0.186} & {\bf 33.26} & 0.853 & {\bf 0.189}   \\
\hline
\end{tabular}
}
\end{center}
\caption{Results of textured single view reconstruction.}
\label{table:svr_numbers}
\end{table}

In addition to FID evaluated on a set of shapes, we use Structural Similarity Index Measure (SSIM)~\cite{wang2004image} and feature-$l_1$~\cite{texture_fields} to evaluate the quality of each rendered view and then average them, as proposed in TF. Table~\ref{table:svr_numbers} reports the results, with AUV-Net outperforming the baseline. We observe that both methods overfitted on ShapeNet chairs, and are unable to recover the correct textures of complex test shapes. This is likely due to the fact that the dataset has significant variation for chairs, but has insufficient training examples. In addition, we find that SSIM, which is not semantics-aware, may not be a good evaluation metric when the results are not very close to the ground truth. 
As shown in Table~\ref{table:svr_numbers}, the SSIM of ours and the baseline are very close, though other metrics show clear differences.

\vspace{-2mm}
\section{Conclusion, Limitations, and Future Work}
\label{sec:future}
\vspace{-1mm}

\begin{figure}[t!]
\begin{center}
\includegraphics[width=0.9\linewidth]{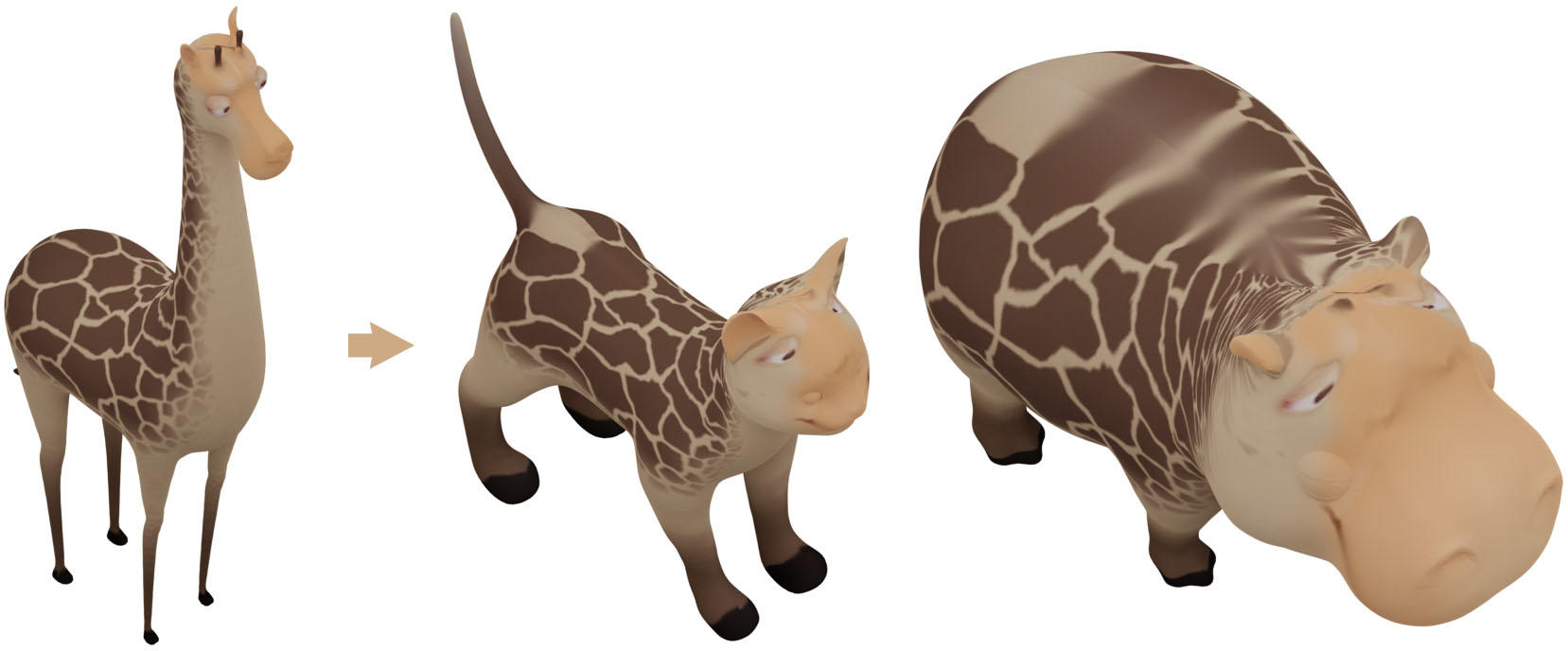}
\end{center}
\caption{
When transferring texture of a cartoon giraffe into other animals, the positions of the eyes are wrong. The cut seam on the hippo is clearly visible, although inpainted.
}
\label{fig:seam}
\end{figure}

We introduce the first method to learn aligned texture maps for a set of shapes in an unsupervised manner. We show that alignment can be done with a simple alignment module inspired by PCA. The resulting texture images of our method are well aligned and disentangled from geometry. They have enabled several applications including texture transfer, texture synthesis, and textured single view 3D reconstruction, which we showcase in our experiments.

There are three main limitations of our approach. First, our method does not handle seams that arise when texturing the shape, and only exploits a shared UV mapper network to alleviate the issue. Therefore, seams may become conspicuous for some shapes after transferring textures, as shown in Fig.~\ref{fig:seam}.
Second, our method does not always find correct correspondences, especially when the textures are messy, e.g., the animal dataset - one can observe that the eyes are not properly aligned in Fig.~\ref{fig:transfer}. Adding weak supervision could help, e.g., annotating two points for the eyes in all training shapes. Third, our method does not handle shapes with complex topology well. In fact, ShapeNet chairs pose a significant difficulty for our method with two basis generators, and we had to use four to avoid overlapping textures. We leave these challenges to future work.

\vspace{-3mm}
\paragraph{Acknowledgements.}
We thank the reviewers for their valuable comments. This work was completed
when the first author was carrying out an internship at NVIDIA.

{\small
\bibliographystyle{ieee_fullname}
\bibliography{egbib}
}

\setcounter{page}{1}
\setcounter{section}{0}

\twocolumn[
\centering
\textbf{\large{AUV-Net: Learning Aligned UV Maps for Texture Transfer and Synthesis}} \\
\vspace{0.5em}\large{Supplementary Material} \\
\vspace{1.0em}
]

This document provides supplementary material for AUV-Net. It contains details regarding network architectures, loss function definitions, training protocols, dataset licenses, and ethics discussions. More qualitative results can be found in \href{https://youtu.be/UTzH8WB-Xl0}{https://youtu.be/UTzH8WB-Xl0}.

\section{Implementation details}

\subsection{Detailed network architectures}

The detailed architectures of sub-modules in our network are shown in Fig.~\ref{fig:detail_net}.

\subsection{Texture transfer for unseen shapes}

Given a new shape not in the original training set, we put 100 duplicates of that shape in the training dataset, and then continue training stage 3 for 200 epochs. During training, we fix the weights of the basis generators to avoid affecting the already well-trained texture basis. If the new shape does not have textures, we remove the color loss. It takes about 3 hours to finish training for the new shape on one NVIDIA RTX 3080 Ti GPU.

\subsection{Image generative model for texture synthesis}

We train a StyleGAN2 ~\cite{Karras2019stylegan2} on the aligned texture images, after stitch multiple texture images of the same object into one single image. For categories with 2 texture images per object, we concatenate the two images horizontally and then resize them to $1024^2$. For chair category with 4 texture images, we put them at the four quadrants of an $1024^2$ image. We further apply Gaussian blur with $\sigma=1$ on the training images to remove small noise, making the edges slightly smoother, and boosting the performance of generation. Training StyleGAN2 takes around 48 hours on 8 Nvidia Tesla v100 GPUs. During inference, we split each generated image back to multiple ones.

\subsection{Textured single view reconstruction}

\begin{figure*}[b!]
\begin{center}
\includegraphics[width=1.0\linewidth]{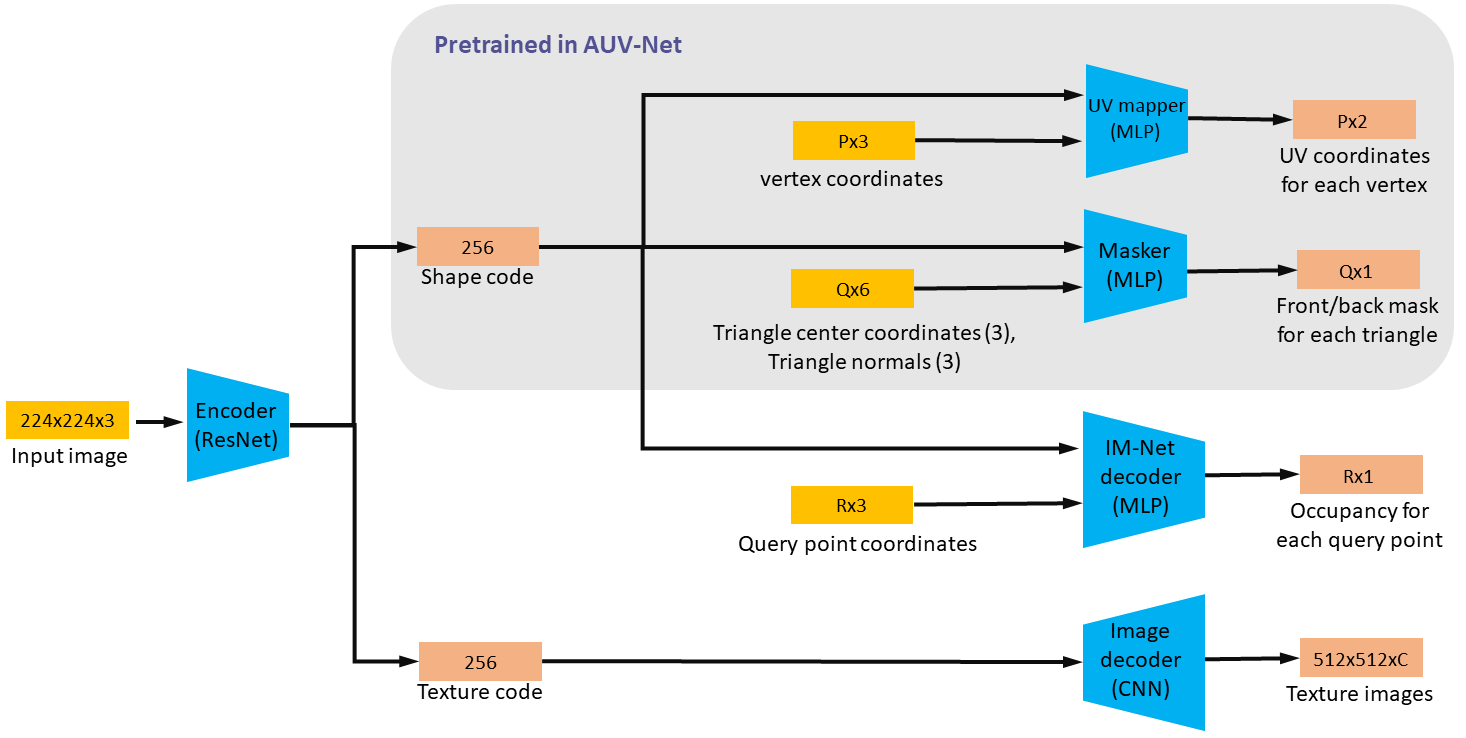}
\end{center}
\caption{ Architecture of network for textured single view reconstruction. Note that the UV mapper and the Masker are from a pretrained AUV-Net trained on the training shapes, and they are used to predict the UV mapping for the predicted shape by the IM-Net decoder. The shape codes and the texture images produced by the pretrained AUV-Net are used in this network as the ground truth.
}
\label{fig:svr_net}
\end{figure*}

The network architecture for this application is shown in Fig.~\ref{fig:svr_net}. The details of the image decoder is shown in Fig.~\ref{fig:detail_net}.  
We follow the implementation of ResNet-18~\cite{resnet} and IM-Net~\cite{imnet} for the image encoder and the implicit shape decoder, respectively. Note that we generate $512^2$ texture images to reduce training time, and the number of output channels (RGB+alpha) is $C=8$ for car category with 2 texture images and $C=16$ for chair category with 4 texture images.

\vspace{-5mm}
\paragraph{Training.}
We first train our AUV-Net on the 3D training set to obtain aligned texture images and the shape codes. Then we use those as ground truth to train the SVR network in Fig.~\ref{fig:svr_net} with MSE losses. Note that the ResNet encoder directly regresses the provided shape codes, therefore the pre-trained UV mapper and Masker are not used during SVR training. The IM-Net decoder also reconstructs shapes directly from the provided shape codes, therefore it can be trained separately without other sub-networks. Pre-training AUV-Net on each ShapeNet category (200 epochs) takes around 36 hours on one NVIDIA RTX 3080 Ti GPU. Training the SVR network (2000 epochs) takes around 36 hours on 8 Nvidia Tesla v100 GPUs.

\vspace{-5mm}
\paragraph{Inference.}
For each input image, the ResNet encoder predicts a shape code and a texture code. We use the IM-Net decoder to obtain the implicit field of the shape from the predicted shape code, and then use Marching Cubes ~\cite{lorensen1987marching} to obtain the triangle mesh of the shape. Then we feed the vertices and the triangles of the shape into the UV mapper and the masker to obtain the UV maps. The texture images are predicted by the image decoder from the texture code.

\section{Configurations for different datasets}

The main differences between the configurations of different datasets are the number of basis generators, basis generator hyper-parameters (the hidden layer size and the number of basis images), the definition of the prior loss, and the weights of the five loss terms. The basis generator hyper-parameters are different because we reduce the sizes of the networks in certain categories to reduce training time. The other hyper-parameters are set differently according to the properties of the datasets.

\subsection{Triplegangers ~\cite{Triplegangers} heads}

\begin{figure*}[b!]
\begin{center}
\includegraphics[width=1.0\linewidth]{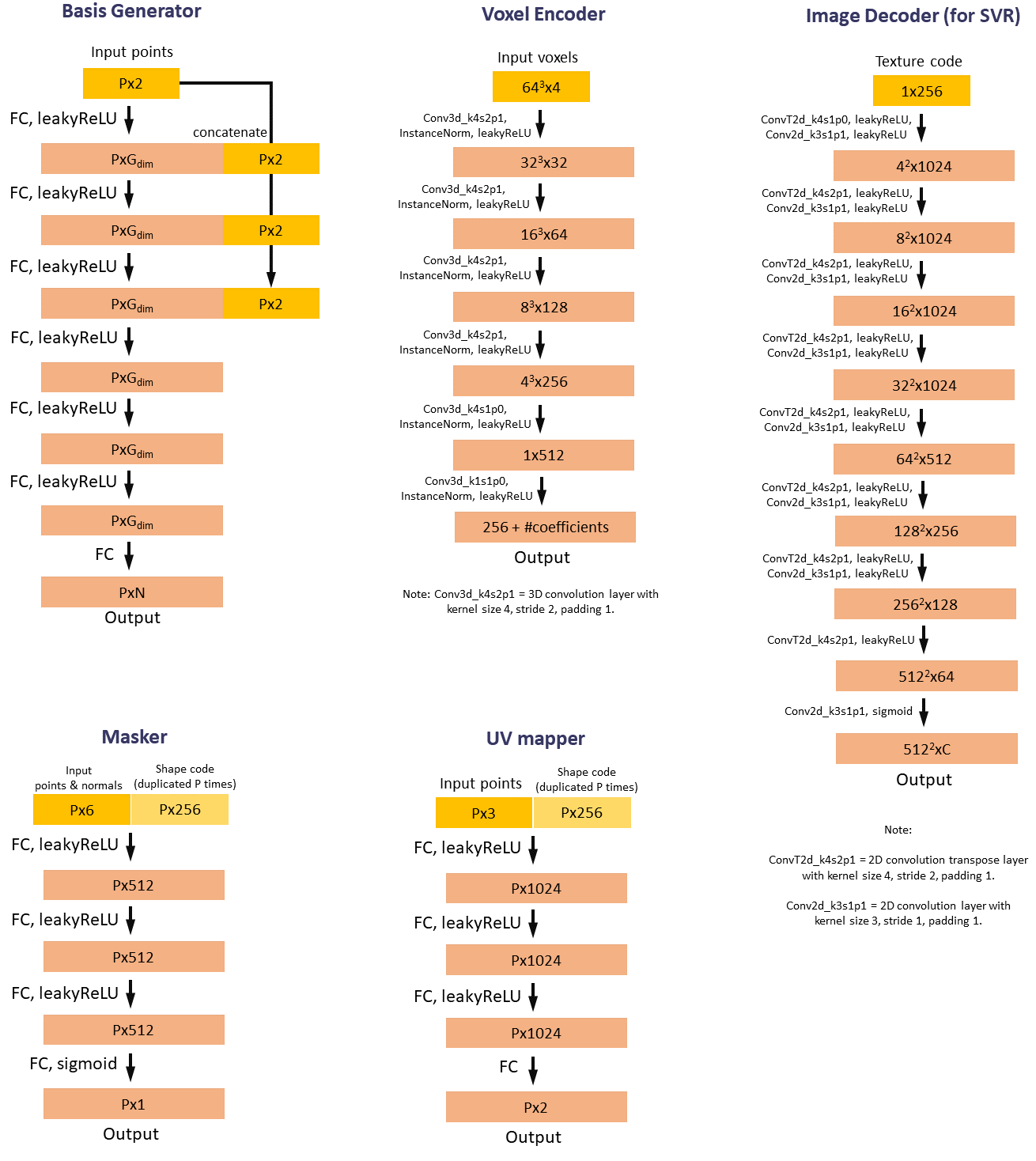}
\end{center}
\caption{ Detailed network architectures.
}
\label{fig:detail_net}
\end{figure*}

We use 2 basis generators. The first basis generator has hidden layer size $G_{dim}=1024$ (see Fig.~\ref{fig:detail_net}), and outputs 64 channels (=64 basis images). The second basis generator has hidden layer size $G_{dim}=128$, and outputs 16 channels. We use different hidden layer sizes and output channels for those basis generators to reduce training time, considering that the second basis generator has significantly less details to represent (see Fig.~\ref{fig:network} in the main paper).

The prior loss is defined as
\begin{equation}
L_{p} = \frac{1}{N} \sum_{i=1}^{N} (m_i - n_i)^2 + (p_i^x - q_i^x)^2 + (p_i^y - q_i^y)^2.
\end{equation}
where $m_i$ is the masking value predicted by the masker for the $i$-th input point $p_i$. $n_i$ is defined as $1$ if the normal of $p_i$ in the z direction is greater than $-0.5$, and $0$ otherwise. (The heads are facing z direction, and the top is y direction.) $p_i^x$ is the x coordinate of $p_i$, $q_i^x$ is the x coordinate of $q_i$.

We train the first stage for 10 epochs, with $\{ w_{c},w_{n},w_{x},w_{s},w_{p} \} = \{ 1,0.5,100,100,1 \}$;  second stage for 2,000 epochs, with $\{ 1,0.5,1,1,0 \}$; third stage for 2,000 epochs, with $\{ 1,0.5,100,100,0 \}$.

\subsection{RenderPeople ~\cite{RenderPeople} human bodies}

We use 2 basis generators. Both basis generators have hidden layer size $G_{dim}=1024$, and output 64 channels.

The prior loss is defined as
\begin{equation}
L_{p} = \frac{1}{N} \sum_{i=1}^{N} (m_i - n_i)^2 + (p_i^x - q_i^x)^2 + (p_i^y - q_i^y)^2.
\end{equation}
where $n_i$ is defined as $1$ if the normal of $p_i$ in the yz direction is greater than $-0.5$, and $0$ otherwise. (The bodies are facing z direction, and the top is y direction.)

We train the first stage for 10 epochs, with $\{ 1,0.5,1,1,1 \}$;  second stage for 2,000 epochs, with $\{ 1,0.1,1,1,0 \}$; third stage for 2,000 epochs, with $\{ 1,0.5,100,100,0 \}$.

\subsection{Turbosquid ~\cite{Turbosquid} animals}

We use 2 basis generators. Both basis generators have hidden layer size $G_{dim}=1024$, and output 64 channels.

The prior loss is defined as
\begin{equation}
L_{p} = \frac{1}{N} \sum_{i=1}^{N} (m_i - n_i)^2 + (p_i^y - q_i^x)^2 + (p_i^z - q_i^y)^2.
\end{equation}
where $n_i$ is defined as $1$ if the normal of $p_i$ in the x direction is greater than $0$, and $0$ otherwise. (The animals are facing z direction, and the top is y direction.)

We train the first stage for 10 epochs, with $\{ 0.1,1,100,100,100 \}$;  second stage for 2,000 epochs, with $\{ 0.1,1,100,100,0 \}$; third stage for 2,000 epochs, with $\{ 0.1,1,100,10,0 \}$.

\subsection{Turbosquid ~\cite{Turbosquid} cars}

We use 2 basis generators. The first basis generator has hidden layer size $G_{dim}=1024$, and outputs 64 channels. The second basis generator has hidden layer size $G_{dim}=128$, and outputs 16 channels.

The prior loss is defined as
\begin{equation}
L_{p} = \frac{1}{N} \sum_{i=1}^{N} (m_i - n_i)^2 + (p_i^x - q_i^x)^2 + (p_i^z - q_i^y)^2.
\end{equation}
where $n_i$ is defined as $1$ if the normal of $p_i$ in the y direction is greater than $-0.5$, and $0$ otherwise. (The cars are facing inverse z direction, and the top is y direction.)

We train the first stage for 200 epochs, with $\{ 1,0.1,100,10,1 \}$;  second stage for 1,800 epochs, with $\{ 1,0.1,1,10,0 \}$; third stage for 2,000 epochs, with $\{ 1,0.1,1000,100,0 \}$.  We set $w_{p}$ to $0$ after 10 epochs in the first stage.

\subsection{ShapeNet ~\cite{chang2015shapenet} cars}

We use 2 basis generators. The first basis generator has hidden layer size $G_{dim}=1024$, and outputs 64 channels. The second basis generator has hidden layer size $G_{dim}=128$, and outputs 16 channels.

The prior loss is defined as
\begin{equation}
L_{p} = \frac{1}{N} \sum_{i=1}^{N} (m_i - n_i)^2 + (p_i^x - q_i^x)^2 + (p_i^z - q_i^y)^2.
\end{equation}
where $n_i$ is defined as $1$ if $p_i^y>0$ or the normal of $p_i$ in the y direction is greater than $-0.5$, and $0$ otherwise. (The cars are facing x direction, and the top is y direction.)

We train the first stage for 20 epochs, with $\{ 1,0.1,10,10,1 \}$;  second stage for 40 epochs, with $\{ 1,0.1,1,10,0 \}$; third stage for 140 epochs, with $\{ 1,1,100,100,0 \}$.  We set $w_{p}$ to $0$ after 5 epochs in the first stage.

\subsection{ShapeNet ~\cite{chang2015shapenet} chairs}

We use 4 basis generators. All basis generators have hidden layer size $G_{dim}=512$, and output 64 channels.

Since there are four texture images, we modify the masker to output four channels. We only input point coordinates and the shape code to the masker, and let it predict the point normal $n^{pred}_i$ (not necessarily unit normal) and the mask $m^{pred}_i$. We then compute the dot product between the predicted normal and the ground truth normal to get a second mask $m^{n}_i = sigmoid(n^{pred}_i \cdot n^{gt}_i)$. The final output masks are
\begin{equation}
\left[ \begin{array}{cc}
m_i^a & m_i^c \\
m_i^b & m_i^d
\end{array} \right] = 
\left[ \begin{array}{c}
m^{n}_i \\
1-m^{n}_i
\end{array} \right] \cdot [m^{pred}_i \;\;\; 1-m^{pred}_i]
\end{equation}
The prior loss is defined as
\begin{align}
\begin{aligned}
L_{p} & = \frac{1}{N} \sum_{i=1}^{N} (m_i^a - s_i^a)^2 + (m_i^b - s_i^b)^2 + (m_i^c - s_i^c)^2 \\
 & + (m_i^d - s_i^d)^2 + (t_i^x - q_i^x)^2 + (t_i^y - q_i^y)^2.
\end{aligned}
\end{align}
where $m_i^a$,$m_i^b$,$m_i^c$,$m_i^d$ are the masking values predicted by the masker. $s_i^a$,$t_i^x$,etc. are defined as follows. The chairs are facing x direction, and the top is y direction. The chairs are normalized in a unit cube centered at origin. $n_i$ is the unit normal of $p_i$. 

\begin{align}
\begin{aligned}
p_{max}^y &= \max_i p_i^y, \\
p_i^y{}' &= p_i^y - p_{max}^y - 0.05, \\
m_i{}' &= \left\{ \begin{array}{ll}
1 & \textrm{if} \;\; p_i^x n_i^x + p_i^y{}' n_i^y + p_i^z n_i^z <0,\\
0 & \textrm{otherwise.}
\end{array} \right. \\
p_{seat}^y &= \max_{ i \in \{ i \; | \; p_i^x>0 \;\land\; p_i^x<0.1 \;\land\; p_i^z>-0.05 \;\land\; p_i^z<0.05 \} } p_i^y \\
m_i{}'' &= \left\{ \begin{array}{ll}
1 & \textrm{if} \;\; p_i^y>p_{seat}^y-0.2,\\
0 & \textrm{otherwise.}
\end{array} \right. \\
\left[ \begin{array}{cc}
s_i^a & s_i^c \\
s_i^b & s_i^d
\end{array} \right] &= 
\left[ \begin{array}{c}
m_i{}' \\
1-m_i{}'
\end{array} \right] \cdot [m_i{}'' \;\;\; 1-m_i{}''] \\
d_i &= \sqrt{ p_i^x{}^2+p_i^z{}^2+4(p_i^y-p_{seat}^y)^2 } \Big{/} \sqrt{ p_i^x{}^2+p_i^z{}^2 } \\
t_i^x, t_i^y &= p_i^x d_i, p_i^z d_i
\end{aligned}
\end{align}

We train the first stage for 50 epochs, with $\{ 1,1,10,100,100 \}$;  second stage for 50 epochs, with $\{ 1,1,10,10,0 \}$; third stage for 100 epochs, with $\{ 1,1,100,100,0 \}$. We set $w_{p}$ to $0$ after 5 epochs in the first stage.

\section{Dataset licenses}
In the paper, we use 3D assets of cars and chairs from ShapeNet~\cite{chang2015shapenet},  animals and cars from Turbosquid~\cite{Turbosquid},  human faces from Triplegangers~\cite{Triplegangers}, human bodies from Renderpeople~\cite{RenderPeople}. The licenses for using Renderpeople, Triplegangers and TurboSquid datasets were obtained through commercial agreements. The license information of ShapeNet is provided on its website \footnote{https://shapenet.org/terms}.  We carefully inspected the datasets we use and did not find identifiable information or offensive content. To run the comparison with baseline methods, we use the source code provided by the authors on GitHub. 

\section{Ethics discussions}

\subsection{Potential negative societal impacts}
We present a generic method for synthesizing textures for  a variety of classes of 3D objects.  It does also handle 3D human face/body shapes and textures.  Our method can create 3D avatars for users by transferring textures to 3D meshes, or reconstructing textured meshes from input photos.   Like any other machine learning models, AUV-NET is prone to biases imparted through training data which requires an abundance of caution when applied to sensitive applications.  It is not recommended in off-the-shelf settings where privacy or erroneous results can lead to potential misuse or any harmful application. For purposes of real deployment, one would need to carefully inspect and de-bias the dataset to depict the target distribution of a wide range of possible lighting conditions, clothing,  skin tones, or at the intersection of race and gender.

\subsection{Personal data/Human subjects}
Our paper uses 3D scans of human faces and bodies obtained from Triplegangers and Renderpeople, respectively.   The data collection and ethics approvals were taken care of by the dataset providers. More information about the dataset can be found on the websites of data providers.


\end{document}